\documentclass[10pt,twocolumn,letterpaper]{article}

\usepackage{cvpr}
\usepackage{times}
\usepackage{epsfig}
\usepackage{graphicx}
\usepackage{amsmath}
\usepackage{amssymb}

\usepackage{comment}

\usepackage{xfrac}
\usepackage{subcaption}
\usepackage[export]{adjustbox}
\usepackage[table]{xcolor}
\usepackage{rotating}
\usepackage{array}
\usepackage{makecell}
\usepackage{color}
\usepackage{booktabs}
\usepackage{tablefootnote}
\usepackage{colortbl}
\usepackage{tabularx}
\usepackage{multirow}

\usepackage[nolist]{acronym}
\usepackage{units}

\makeatletter
\@namedef{ver@everyshi.sty}{}
\makeatother
\usepackage{pgfplots}
\usepackage{tikz} 
\pgfplotsset{compat=newest}
\usepgfplotslibrary{groupplots}
\usetikzlibrary{matrix}

\definecolor{myblue}{rgb}{.1,0.3,0.9}
\definecolor{rowblue}{RGB}{220,230,240}

\usepackage[pagebackref=true,breaklinks=true,letterpaper=true,colorlinks,bookmarks=false]{hyperref}

\cvprfinalcopy 

\ifcvprfinal\fi
\begin{document}


\title{Gated3D: Monocular 3D Object Detection\\From Temporal Illumination Cues} 

\author{
\hspace{-0.12in}Frank Julca-Aguilar$^{1}$ \hspace{0.04in}
Jason Taylor $^{1}$ \hspace{0.04in}
Mario Bijelic $^{2,3}$ \hspace{0.04in} 
Fahim Mannan $^{1}$ \hspace{0.04in} 
Ethan Tseng $^{4}$ \hspace{0.04in}
Felix Heide$^{1,4}$ \vspace{8pt}
\\
\textsuperscript{1}Algolux\hspace{0.15in}
\textsuperscript{2}Daimler AG\hspace{0.15in}
\textsuperscript{3}Ulm University\hspace{0.15in}
\textsuperscript{4}Princeton University\hspace{0.15in}
}
\maketitle


\definecolor{Gray}{rgb}{0.5,0.5,0.5}
\definecolor{darkblue}{rgb}{0,0,0.7}
\definecolor{orange}{rgb}{1,.5,0} 
\definecolor{red}{rgb}{1,0,0} 

\definecolor{dai_ligth_grey}{RGB}{230,230,230}
\definecolor{dai_ligth_grey20K}{RGB}{200,200,200}
\definecolor{dai_ligth_grey40K}{RGB}{158,158,158}
\definecolor{dai_ligth_grey60K}{RGB}{112,112,112}
\definecolor{dai_ligth_grey80K}{RGB}{68,68,68}
\definecolor{dai_petrol}{RGB}{0,103,127}
\definecolor{dai_petrol20K}{RGB}{0,86,106}
\definecolor{dai_petrol40K}{RGB}{0,67,85}
\definecolor{dai_petrol80}{RGB}{0,122,147}
\definecolor{dai_petrol60}{RGB}{80,151,171}
\definecolor{dai_petrol40}{RGB}{121,174,191}
\definecolor{dai_petrol20}{RGB}{166,202,216}
\definecolor{dai_deepred}{RGB}{113,24,12}
\definecolor{dai_deepred20K}{RGB}{90,19,10}
\definecolor{dai_deepred40K}{RGB}{68,14,7}
\definecolor{rot}{RGB}{238, 28 35}
\definecolor{apfelgruen}{RGB}{140, 198, 62}
\definecolor{orange}{RGB}{244, 111, 33}
\definecolor{pink}{RGB}{237, 0, 140}
\definecolor{lila}{RGB}{128, 10, 145}
\definecolor{anthrazit}{RGB}{19, 31, 31}

\newcommand{\heading}[1]{\noindent\textbf{#1}}
\newcommand{\todo}[1]{{\textcolor{red}{\bf{TODO: #1}}}}
\newcommand{\comments}[1]{{\em{\textcolor{orange}{#1}}}}
\newcommand{\changed}[1]{#1}
\newcommand{\place}[1]{ \begin{itemize}\item\textcolor{darkblue}{#1}\end{itemize}}
\newcommand{\de}{\mathrm{d}}

\newcommand{\normlzd}[1]{{#1}^{\textrm{aligned}}}

\newcommand{\ttime}{\tau}               
\newcommand{\x}{\Vect{x}}               
\newcommand{\z}{z}               

\newcommand{\npixels}{n}               
\newcommand{\ntime}{t}               

\newcommand{\illfunc}     {g}
\newcommand{\pathfunc}     {s}
\newcommand{\camfunc}     {f}

\newcommand{\irradiance}{E}
\newcommand{\exposure}{b}
\newcommand{\pmdfunc}{f}                
\newcommand{\lightfunc}{g}              
\newcommand{\period}{T}                 
\newcommand{\freqm}{\omega}                
\newcommand{\illphase}{\rho}             
\newcommand{\sensphase}{\psi}             
\newcommand{\pmdphase}{\phi}            
\newcommand{\omphi}{{\omega,\phi}}      
\newcommand{\numperiod}{N}              
\newcommand{\att}{\alpha}               
\newcommand{\pathspace}{{\mathcal{P}}}  

\newcommand{\atan}{\operatorname{atan}}

\newcommand{\Fourier}{\mathfrak{{F}}}         
\newcommand{\conv}     {\otimes}
\newcommand{\corr}     {\star}
\newcommand{\Mat}[1]    {{\ensuremath{\mathbf{\uppercase{#1}}}}} 
\newcommand{\Vect}[1]   {{\ensuremath{\mathbf{\lowercase{#1}}}}} 
\newcommand{\Id}				{\mathbb{I}} 
\newcommand{\Diag}[1] 	{\operatorname{diag}\left({ #1 }\right)} 
\newcommand{\Opt}[1] 	  {{#1}_{\text{opt}}} 
\newcommand{\CC}[1]			{{#1}^{*}} 
\newcommand{\Op}[1]     {\Mat{#1}} 
\newcommand{\minimize}[1] {\underset{{#1}}{\operatorname{argmin}} \: \: } 
\newcommand{\maximize}[1] {\underset{{#1}}{\operatorname{argmax}} \: \: } 
\newcommand{\grad}      {\nabla}

\newcommand{\Basis}{\Mat{H}}         		
\newcommand{\Corr}{\Mat{C}}             
\newcommand{\correlem}{\bold{c}}             
\newcommand{\meas}{\Vect{b}}            
\newcommand{\Meas}{\Mat{B}}            
\newcommand{\MeasNormalized}{\Mat{B}^{\textrm{new}}}            
\newcommand{\Img}{H}                    
\newcommand{\img}{\Vect{h}}             
\newcommand{\latentresponse}{\alpha}

\newenvironment{customlegend}[1][]{%
        \begingroup
        \csname pgfplots@init@cleared@structures\endcsname
        \pgfplotsset{#1}%
    }{%
        \csname pgfplots@createlegend\endcsname
        \endgroup
    }%

    \def\addlegendimage{\csname pgfplots@addlegendimage\endcsname}

   \newcommand{\arcsec}[1]{$\text{#1}"$}

\begin{abstract}
Today's state-of-the-art methods for 
3D object detection are based on lidar, 
stereo, or monocular cameras. Lidar-based
methods achieve the best accuracy, but have a large footprint, high cost, and mechanically-limited angular sampling rates, 
resulting in low spatial resolution at long ranges.
Recent approaches based on low-cost monocular or stereo
cameras promise to overcome these limitations  
but struggle in low-light or low-contrast regions as they rely on passive CMOS sensors.
In this work, we propose a novel 3D object detection 
modality that exploits temporal 
illumination cues from a low-cost monocular gated imager. 
We propose a novel deep detector architecture,
\textit{Gated3D}, that is tailored to  
temporal illumination cues from three gated images. Gated images allow us to exploit  mature 2D object feature extractors
that guide the 3D predictions through a frustum segment estimation. 
We assess the proposed method on a novel 3D detection
dataset that includes gated imagery captured in over 10,000~km of driving data. 
We validate that our method 
outperforms state-of-the-art monocular and stereo 
approaches at long distances. We will release our 
code and dataset, opening up a new sensor modality as an avenue to replace lidar in autonomous driving.

\end{abstract}

\section{Introduction}
\label{sec:introduction}
3D object detection is a fundamental vision task in robotics and autonomous driving. Accurate 3D detections are critical for safe trajectory planning, with applications emerging across disciplines such as autonomous drones, assistive and health robotics, as well as warehouse and delivery robots. RGB-D cameras using correlation time-of-flight~\cite{hansard2012time,kolb2010time,lange00tof}, such as Microsoft's Kinect One, enable robust 3D detection indoors~\cite{song2014sliding,song2016deep} for small ranges. In the past, autonomous driving, which requires long ranges and high depth accuracy, has relied on scanning lidar for 3D detection~\cite{schwarz2010lidar,wang2015voting,engelcke2017vote3deep,yang2018pixor,li20173d,chen2017multi,zhou2018voxelnet,ku2018joint,lang2019pointpillars}. However, while lidar provides accurate depth, existing systems are fundamentally limited by point-by-point acquisition, resulting in spatial resolution that falls off quadratically with distance and linearly with framerate. In contrast to conventional cameras, lidar systems are three orders of magnitude more expensive, suffer from low resolution at long distances, and fail in the presence of strong back-scatter, e.g. in snow or fog~\cite{bijelic2020seeing}. 
%
\begin{figure*}[t!]
	\centering
	\includegraphics[width=\linewidth]{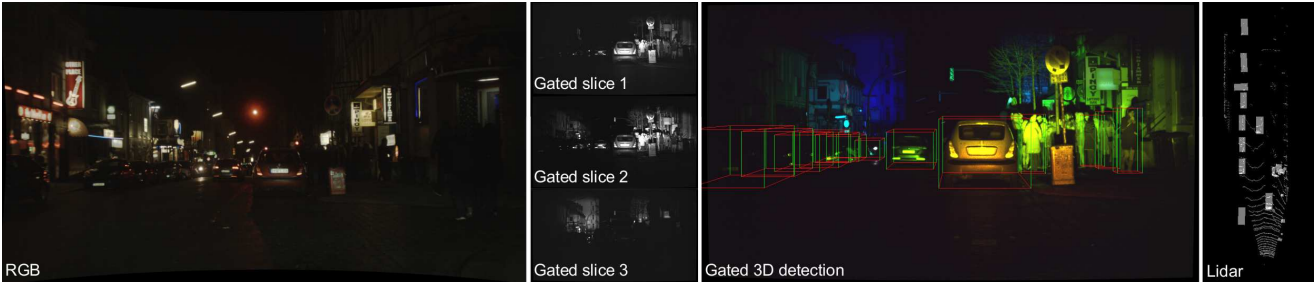}
	\caption{We propose a novel 3D object detection method, which we dub ``Gated3D'', using a flood-illuminated gated camera. The high-resolution of gated images enables semantic understanding at long ranges. In the figure, our gated slices are color-coded with red for slice 1, green for slice 2 and blue for slice 3. We evaluate Gated3D on real data, collected with a scanning lidar Velodyne HDL64-S3D as reference, see overlay on the right.  
	}
	\label{fig:Teaser}
\end{figure*}

Promising to overcome these challenges, a recent line of work proposed \emph{pseudo-lidar sensing}~\cite{wang2019pseudo}, which rely on low-cost sensors, such as stereo~\cite{chen20173d, Chang2018, Kendall2017} or monocular~\cite{chen2016monocular,Godard2017,eigen2014depth} to recover dense depth maps from conventional intensity imagers. Point-clouds are sampled from the depth maps and ingested by 3D detection methods that operate on point-cloud representations~\cite{lang2019pointpillars, zhou2018voxelnet}. More recent methods predict 3D boxes directly from the passive input images~\cite{licvpr2019, brazil2019m3d, simonelli2019disentangling}. Although all of these methods promise low-cost 3D detection with the potential to replace lidar, they rely on \emph{passive} camera-only sensing. Passive stereo approaches degrade at long ranges, where disparities are small, and in low-light scenarios, e.g. at night, when stereo or monocular depth cues are less visible.

In this work, we introduce the first 3D object detection method using gated imaging and evaluate this as a low-cost detection method for long ranges, outperforming recent monocular and stereo detection methods. Similar to passive approaches, we use CMOS sensors but add active temporal illumination. The proposed gated imager captures illumination distributed in three wide gates ($>$ 30~m) for all sensor pixels. Gated imaging~\cite{Heckman1967,Busck2005,Andersson2006,Xinwei2013,schober2017dynamic,adam2017bayesian,gated2depth} allows us to capture several dense high-resolution images distributed continuously across the distances in their respective temporal bin. Additionally, back-scatter can be removed by the the distribution of early gates. Whereas scanning lidar trades off temporal resolution with spatial resolution and SNR, the sequential acquisition of gated cameras trades off dense spatial resolution and SNR (i.e. wide gates) with coarse temporal resolution. We demonstrate that the temporal illumination variations in gated images are a depth cue naturally suited for 3D object detection, without the need to first recover intermediate proxy depth maps~\cite{gated2depth}. Operating on 2D gated slices allows us to leverage existing 2D object detection architectures to guide the 3D object detection task with a novel 
frustum segmentation. The proposed architecture further 
exploits gated images by disentangling the 
semantic contextual features from depth cues in the gates
through a two stream feature extraction.
Relying on the resulting high-resolution 2D feature stacks, the method outperforms existing methods especially at long ranges. The method runs at real-time frame rates and outperforms existing passive imaging methods, independent of the ambient illumination -- promising low-cost CMOS sensors for 3D object detection in diverse automotive scenarios.


Specifically, we make the following contributions:
\begin{itemize}
	\setlength\itemsep{.2em}
	\item We formulate the 3D object detection problem as 
a regression from a frustum segment, computed using 
2D detection priors and the object dimension statistics. 

	\item We propose a novel end-to-end deep neural network architecture that 
	solves the regression problem by effectively integrating 
	depth cues and semantic features from gated images, without 
	generating intermediate depth maps.

	\item We validate the proposed method on real-world driving data acquired with a prototype system in challenging automotive scenarios. We show that the proposed approach detects objects with high accuracy beyond 80~m, outperforming existing monocular, 
	stereo and pseudo-lidar low-cost methods.
	\item We provide a novel annotated 3D gated dataset, covering over 10,000~km driving throughout northern Europe, along with all code.
\end{itemize}

As an example, Figure~\ref{fig:Teaser} shows experimental results of the proposed method. The gated image contains dense information on objects further away in the scene. The advantage of gated sensors for nighttime scenes is also demonstrated in this example, where the pedestrians are not clearly visible in the RGB image.

\section{Related Work}
\vspace{-6pt}
\label{sec:related_work}
\vspace{0.5em}\noindent\textbf{Depth Sensing and Estimation.}
%
Passive acquisition methods for recovering depth from conventional intensity images operate on single monocular images \cite{Chen2018b,Godard2017,Kuznietsov2017,eigen2014depth,saxena2006learning,brazil2019m3d}, temporal sequences of monocular images~\cite{koenderink1991affine,torr1999feature,Ummenhofer2017,Zhou2017}, or on multi-view stereo images~\cite{hartley2003multiple,seitz2006comparison,Chang2018,pilzer2018unsupervised,licvpr2019}. These methods all suffer in low-light and low-contrast scenes. Active depth sensing overcomes these limitations by actively illuminating the scene, and scanning lidar~\cite{schwarz2010lidar} has emerged as an essential depth sensor for autonomous driving, independent of ambient lighting. However, the spatial resolution of lidar is fundamentally limited by the sequential point-by-point scanning frame rate and the sensor cost is significantly higher.
Recently, gated cameras were proposed as an alternative for dense depth estimation~\cite{gated2depth}. Although promising depth estimates have been demonstrated with gated cameras, local artefacts and low-confidence regions in outputs from Gruber et al.~\cite{gated2depth} call into question if its performance for high-quality scene understanding tasks could surpass that of recent monocular and stereo-based methods -- a gap addressed in this work in an end-to-end fashion by directly processing the gated input slices.

\vspace{0.5em}\noindent\textbf{CNN 2D Object Detection.}
Convolutional neural networks (CNNs) for efficient 2D object detection have outperformed classical methods that rely on hand-crafted features by a large margin \cite{ren2015faster}.
The key concept behind such learned object detectors is the classification of image patches at varying positions and scales \cite{sermanet2013overfeat}. Discretized grid cells and predefined object templates (anchor boxes) are regressed and classified by fully-convolutional network architectures \cite{long2015fully}. To this end, two popular directions of research have been explored: single-stage \cite{liu2016ssd,redmon2016you,huang2015densebox,lin2017focal} and proposal-based two-stage detectors \cite{girshick2014rich,girshick2015fast,ren2015faster}.
Two-stage approaches such as R-CNN~\cite{girshick2014rich} and Faster R-CNN~\cite{ren2015faster} generate region proposals for objects in the first stage followed by object classification and bounding box refinement in the second stage \cite{girshick2014rich}.
Single-stage detectors such as SSD \cite{liu2016ssd} and YOLO \cite{redmon2016you} directly predict the final detections and are usually faster than two-stage detectors but with lower accuracy. 
Recently, RetinaNet \cite{lin2017focal} proposed a focal loss that effectively down-weights easily-classified background examples and showed that single-stage detectors trained with this loss can outperform two-stage detectors in terms of accuracy.


\vspace{0.5em}\noindent\textbf{3D Object Detection.}
A large body of work on 3D object detection has explored different scene and measurement representations.
For lidar point cloud data, one direction is to rely on voxel-based representations~\cite{wang2015voting,engelcke2017vote3deep,zhou2018voxelnet,chen2019fast,shi2019pv}.
Unfortunately, the computational cost of the 3D convolutions required for voxel-based approaches is prohibitive for real-time processing~\cite{wang2015voting,engelcke2017vote3deep}.
Alternatively, the height dimension of the voxel grid can be collapsed into feature channels with 2D convolutions performed in the BEV plane~\cite{yang2018pixor,lang2019pointpillars,luo2018furious}, trading off height information for computational efficiency.


Although current state of the art relies on lidar, recent work has been attempting to close the performance gap with low-cost passive sensors due to the limitations of scanning lidar, such as cost, size, low angular resolution and failure in strong back-scatter.

Earlier work on monocular \cite{chen2016monocular, simonelli2019disentangling, brazil2019m3d} and stereo \cite{licvpr2019} methods leveraged convolutional architectures from 2D object detection, extracting depth information from stereo disparity cues or geometric constraints in an end-to-end fashion.
More recently, pseudo-lidar~\cite{wang2019pseudo} showed that point cloud input representations can be used with passive imaging approaches by first estimating depth maps. Several methods have since followed this approach with monocular \cite{weng2019monocular, ma2019accurate} and stereo \cite{you2019pseudo} depth estimation.
PatchNet~\cite{ma2020rethinking} proposed that the advantage of pseudo-lidar is its explicit depth information in its input rather than the point cloud representation. Instead, PatchNet uses a 2D convolutional architecture with the estimated (x,y,z) coordinates of each pixel as its input. Estimating the depth prior to the detection network effectively disentangles depth information from object appearance, improving the detection accuracy.

In this work, we propose a method for 3D detection using 2D gated images, offering a low-cost solution comparable to passive sensors with improved detection accuracy. This input representation allows us to leverage the rich body of efficient 2D convolutional architectures for the task of 3D object detection, while the gated slices represent depth more effectively than RGB images.

\section{Gated Imaging}
\label{sec:gated_imaging}
%

Gated imaging is an emerging sensor technology for self-driving cars which relies on active flash illumination to allow for low-light imaging (e.g. night driving) while reducing back-scatter in adverse weather situations such as snow or fog~\cite{gated2depth}.

\begin{figure}[t!]
\centering
\resizebox{0.90\linewidth}{!}{
    \begin{tikzpicture}
    \tikzstyle{box} = [draw,very thick,rounded corners=.1cm,inner sep=5pt,minimum height=3em, text width=5em, align=center] 
    \tikzstyle{circ} = [draw=green, very thick,circle,minimum size=1.5cm] 
    
    \node (laser_text) at (-1,0.95) {Pulsed Laser};    
    \node (cam_text) at (-1,3.8) {Gated Sensor};  
    \node (cam) at (-1,2.6) {\includegraphics[height=1.8cm]{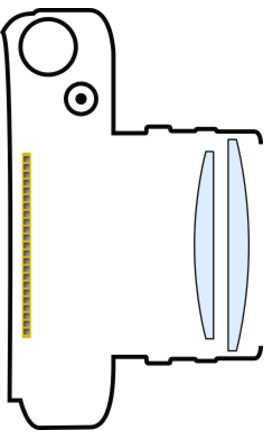}};  
    \node (laser) at (-0.5,1.5) {\includegraphics[height=0.6cm]{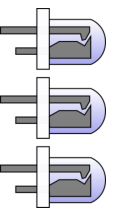}};
    
    \coordinate (v0) at (-2.25,2.0) {};
    \draw [very thick, ->]  (v0) |- (laser);
    \draw [very thick, ->]  (v0) |- (cam);

\begin{axis}
	             		    					[
	             		    					title=Range-Intensity Profile,
	             		    					title style={yshift=-2mm,},
	             		    					xlabel=Distance $r$ \unit{[m]},
	             		    					ylabel=$C\left( r \right)$,
	             		    					ytick=\empty,
	             		    					yticklabels={,,},
	             		    					xmin=0, xmax=100,
	             		    					ymin=0, ymax=9000,
	             		    					height=4.0cm,
	             		    					width=10.0cm,
	             		    					xshift=0.8cm,
	             		    					yshift=1.2cm,
	             		    					xtick style={draw=none},
	             		    					]
	             		    				
	             		    				\addplot+ 	[
	             		    							very thick,
	             		    							solid, 
	             		    							dai_ligth_grey40K,
	             		    							mark=none
	             		    							]
	             		    				table[x index=0,y index=3,col sep=space]{fig/image_formation/bwv_rip.txt};
	             		    				
	             		    				\addplot+ 	[
	             		    							very thick,
	             		    							solid, 
	             		    							dai_petrol,
	             		    							mark=none
	             		    							]
	             		    				table[x index=0,y index=2,col sep=space]{fig/image_formation/bwv_rip.txt};
	             		    				
	             		    				\addplot+ 	[
	             		    							very thick,
	             		    							solid, 
	             		    							dai_deepred,
	             		    							mark=none
	             		    							]
	             		    				table[x index=0,y index=1,col sep=space]{fig/image_formation/bwv_rip.txt};
	             		    				
	             		    				\end{axis}

\node (pedestrian) at (6,1.65) {\includegraphics[height=0.85cm]{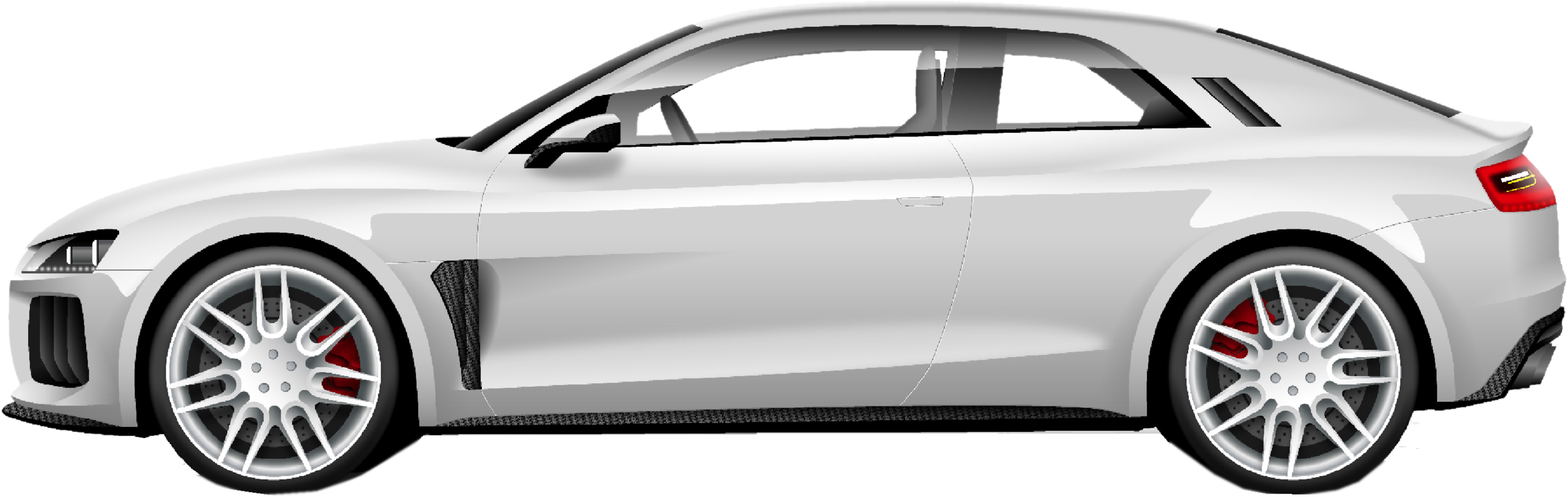}};

\node[text width=3.8cm,align=center] (slice1) at (-0.5,5.45) {Gated Slice 1 \vspace{1mm}\\ \includegraphics[width=3.8cm]{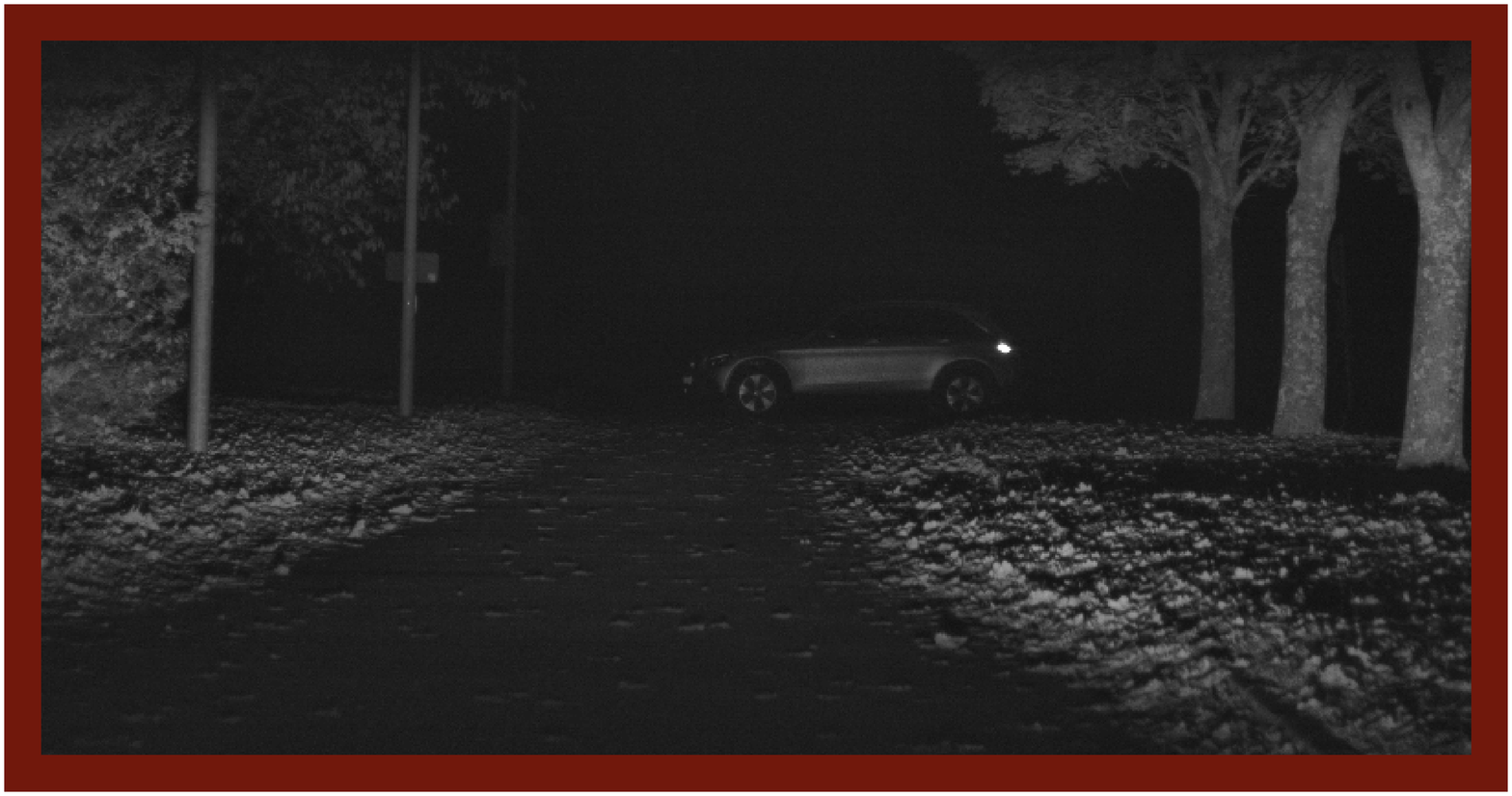}}; 

\node[text width=3.8cm,align=center] (slice2) at (3.5,5.45) {Gated Slice 2 \vspace{1mm}\\ \includegraphics[width=3.8cm]{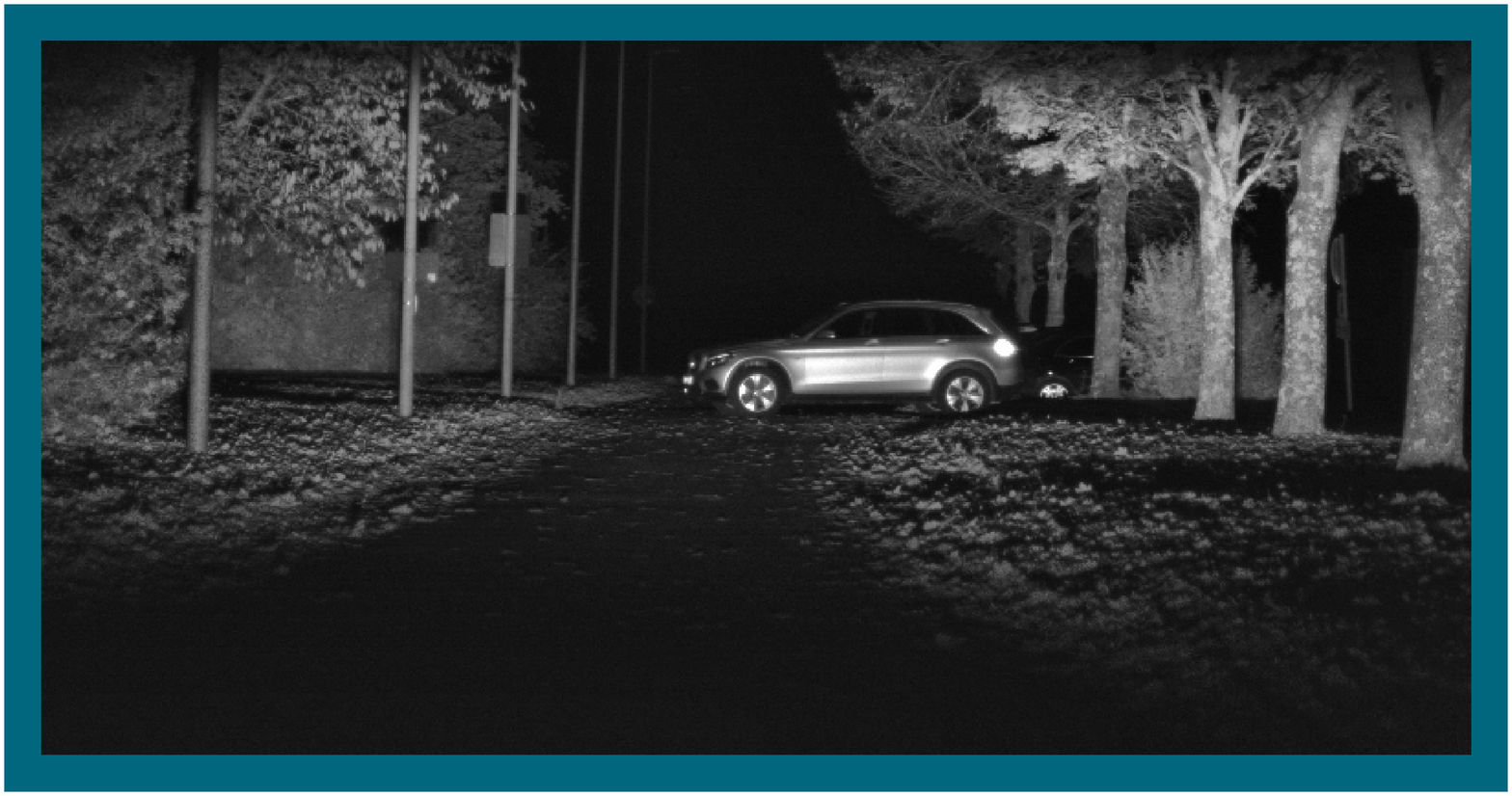}}; 

\node[text width=3.8cm,align=center] (slice3) at (7.5,5.45) {Gated Slice 3 \vspace{1mm}\\ \includegraphics[width=3.8cm]{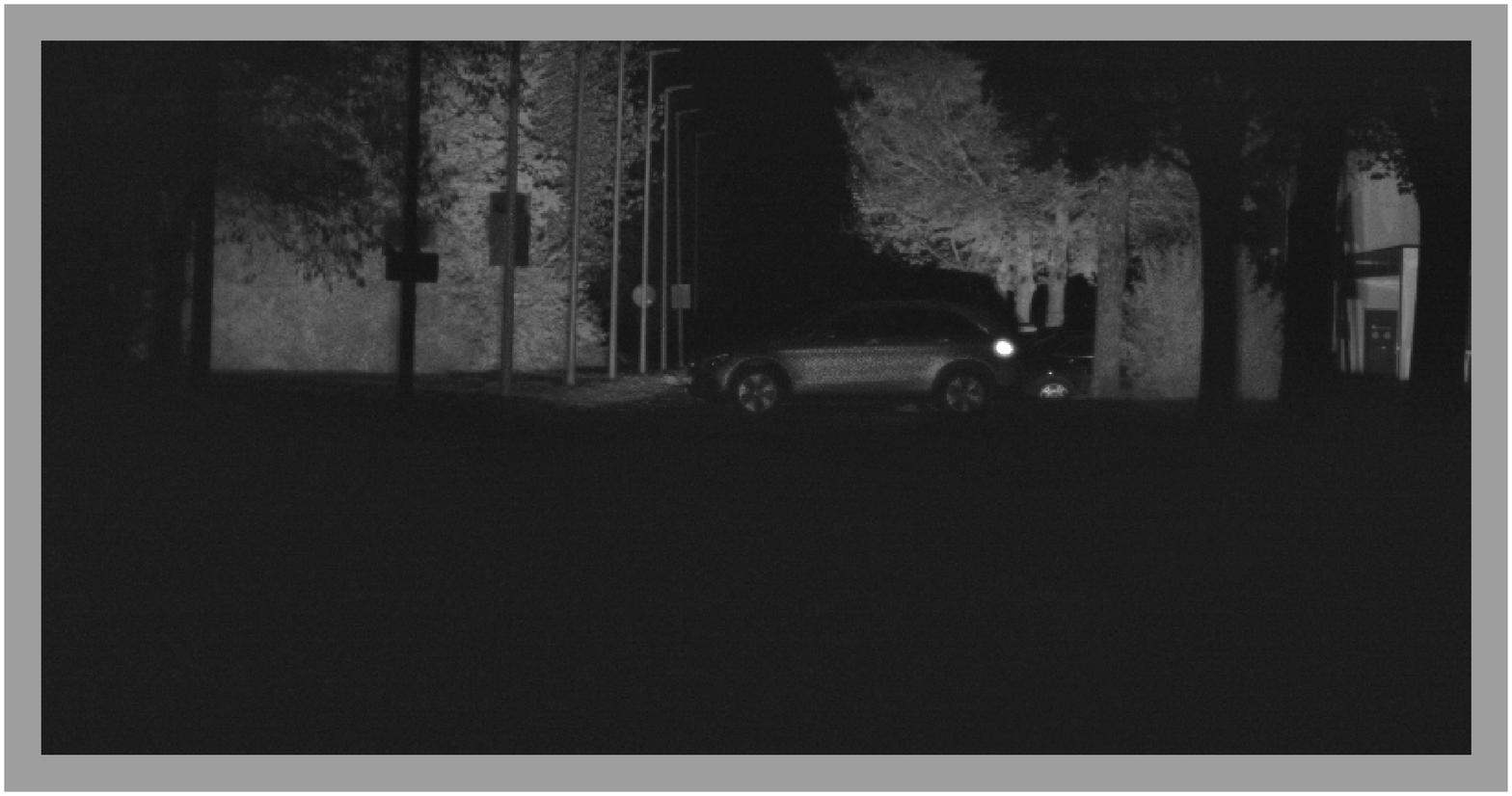}}; 
   
    \end{tikzpicture}}
    \vspace*{-4pt}
        \caption{A gated system consists of a pulsed laser source and a gated imager that are time-synchronized. The range-intensity profile (RIP) $C_i(r)$ describes the distance-dependent illumination for a slice $i$. A car at a certain distance appears with a different intensity in each slice according to the RIP.}
        \label{fig:image_formation}
    \vspace*{-2pt}
\end{figure}
As shown in Figure~\ref{fig:image_formation}, a gated imaging system consists of a flood-illuminator and synchronized gated image sensor that integrates photons falling in a window of round-trip path-length $\xi c$, where $\xi$ is a delay in the gated sensor and $c$ is the speed of light.
Following \cite{gated2depth}, the range-intensity profile (RIP) $C(r)$ describes the distance-dependent integration, which is independent of the scene and given by 
\vspace{-4pt}
\begin{equation}
C\left( r \right) = \int\limits_{-\infty}^{\infty} g\left( t - \xi \right) p\left(t - \frac{2r}{c}\right) \beta\left( r \right) \textrm{d}t, \label{eq:gating_equation}
\end{equation}
where $g$ is the temporally modulated camera gate, $p$ the laser pulse profile and $\beta$ models atmospheric interactions. 
Assuming now a scene with dominating lambertian reflector with albedo $\alpha$ at distance $\tilde{r}$, the measurement for each pixel location is obtained by
\vspace{-4pt}
\begin{equation}
z = \alpha C(\tilde{r})  + \eta_\text{p} \left( \alpha C(\tilde{r}) \right) + \eta_\text{g} ,
\label{eq:noise_model}
\end{equation}
where $\eta_\text{p}$ describes the Poissonian photon shot noise and $\eta_\text{g}$ the Gaussian read-out noise \cite{Foi2008}.
In this work, we capture three images $\mathbf{Z}_i \in \mathbb{N}^{\text{height} \times \text{width}}$ for $i \in \{1,2,3\}$ with different profiles $C_i(r)$ that intrinsically encode depth into the three slices.

\section{3D Object Detection from Gated Images} 
\label{sec:method}
In this section, we introduce \emph{Gated3D}, a novel model
for detecting 3D objects from temporal illumination cues in gated images. 
Given three gated images, the proposed network determines the 3D location, dimensions, orientation and class of the objects in the scene.



\begin{figure*}[t!]
	\centering
	\includegraphics[trim=115 167 75 42, clip, width=\linewidth]{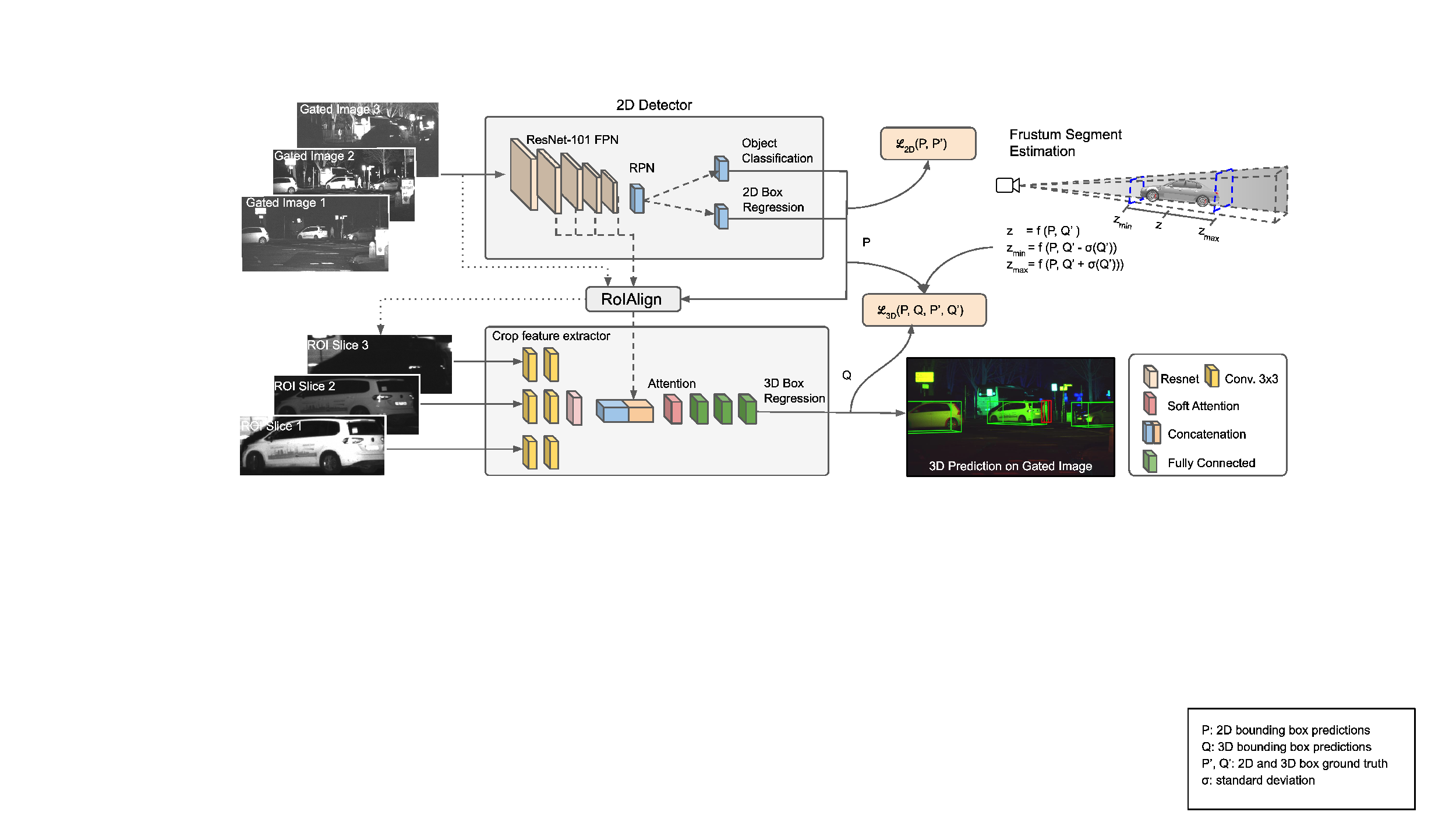}
		\vspace*{-15pt}
	\caption{From three gated slices, the proposed \emph{Gated3D} architecture detects objects and predicts their 3D location, dimension and orientation. 
	Our network employs a 2D detection network 
	to detect ROIs. The resulting 2D boxes are used to crop regions from both the backbone network and input gated slices. 
	Our 3D network estimates the 3D object parameters using a frustum segment 
	computed from the 2D boxes and 3D statistics of the training data. The network processes the gated slices separately, then fuses 
	the resulting features with the backbone features and estimates the 3D bounding box parameters.}
	\label{fig:architecture}
		\vspace*{-4pt}
\end{figure*}

\paragraph{Architecture Overview}
The proposed architecture is illustrated in Figure~\ref{fig:architecture}.
Our model is composed of a 2D detection network, 
based on Mask R-CNN~\cite{He:2017}, and a
3D detection network designed to effectively integrate semantic, contextual, and depth information from gated images. 
The model is trained end-to-end using only 3D bounding box annotations with no additional depth supervision.

The 2D detector predicts bounding boxes that guide the feature extraction with a 
FPN~\cite{Lin:2017} backbone. 
These 2D boxes are used to estimate frustum segments that constrain the 3D location.
In addition to these geometric estimates, the 3D detection network receives the cropped and resized regions of interest extracted from both the input gated slices and the backbone features.
To extract contextual, semantic and depth information from the temporal intensity variations of the gated images, our 3D detection network applies two separate convolution streams: one for the backbone features and another for the gated input slices. 
The resulting features are fed into a sequence of fully-connected layers that predict the 3D location, dimensions, and orientation of the objects.

The remainder of this section details our proposed 2D object detection network~\ref{sec:architecture2d}, 
3D prediction network architecture~\ref{sec:architecture3d} and the loss
functions for training~\ref{sec:loss}.





\subsection{2D Object Detection Network}
\label{sec:architecture2d}
The proposed 2D detection network uses a FPN~\cite{Lin:2017} as a backbone 
and RoIAlign for extracting crops of both the features and input gated slices. 
We extract features maps $P_{2}, P_{3}, P_{4}$ 
and $P_{5}$ of the backbone, as defined in~\cite{Lin:2017}. 

Our 2D object detection network follows a two-stage architecture, 
where the final 2D box detections are refined from proposals output by 
a region proposal network (RPN). 
In contrast to Mask RCNN~\cite{He:2017}, we use these 2D detections instead of the RPN proposals for 3D detection.
Using the refined 2D detections allows the 3D box prediction network to obtain more precise region features, especially from the input gated slices, and a more precise frustum segment, which is essential for depth estimation.

%

\subsection{3D Object Detection Network}
\label{sec:architecture3d}
Our 3D prediction network fuses the extracted features from both the input gated slices and the backbone features.
The gated stream extracts depth cues from the cropped gated input slices with a sequence of convolutions per slice, without parameter sharing.
These convolutions consist of three layers with $3\times3\times16$, $3 \times 3 \times32$ and $3 \times 3 \times32$ kernels.
The network fuses the three gated features and the backbone features by concatenating along the channel dimension and processing with 5 residual layers. Instead of pooling or flattening the resulting features, an attention subnetwork produces softmax attention maps for each feature channel which are used for a weighted sum over the height and width of the features. The resulting feature vectors are fed into two fully connected layers, followed by a final layer that generates eight 3D bounding box coefficients. 

\begin{figure}[t!]
\vspace{-8pt}
	\centering
 	\includegraphics[width=\columnwidth]{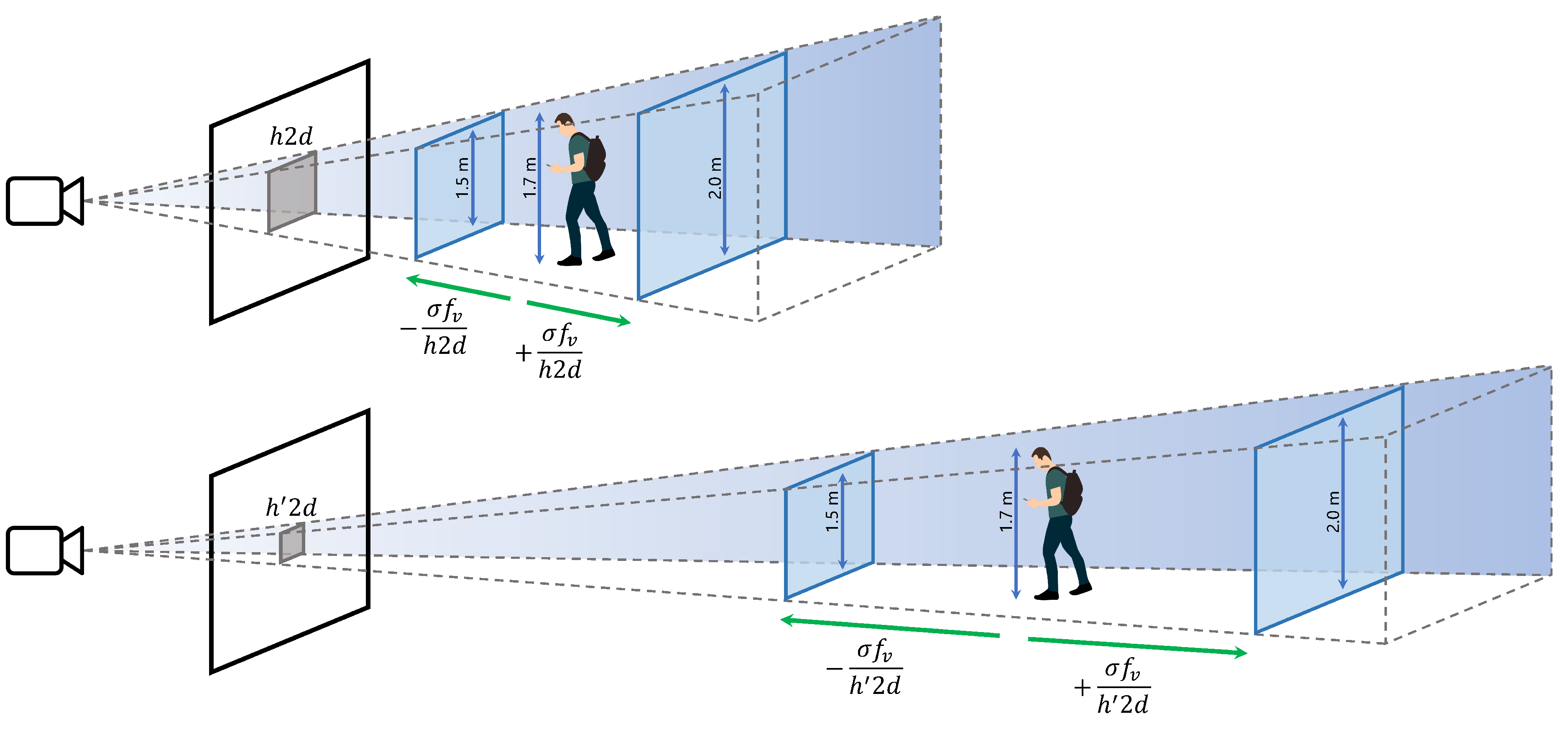}
	\caption{There is an infinite number of 3D cuboids that can project to a given bounding box $P$. However, the object location can be reasonably estimated using the object height, 
	its projected height, and the vertical focal length.} 
	\label{fig:depth_estimation}
	\vspace*{-4pt}
\end{figure}

We denote an object's predicted 2D bounding box 
as $P=(c, u, v, h_{u}, w_{v})$, where $c$ is object's class, 
$(u, v)$ is the bounding box center, 
and $(h_{u}, w_{v})$ define its height and width, respectively. 
The 3D detection network takes $P$ and estimates a set of parameters $Q$, 
that define a 3D bounding box whose projection is given by $P$.
The problem of estimating $Q$ is ill-posed as given a 
specific 2D bounding box $P$, there are an infinite number of 
3D boxes that can be projected to $P$. However, 
we can restrict the range of locations 
of $Q$ to a segment of the 3D viewing frustum extracted from $P$,
using the object's approximate dimensions and $P$. 
See Figure~\ref{fig:depth_estimation} 
for an illustration.

Estimating the 3D location is aided by restricting the 
object's location to a specific frustum region 
similar to~\cite{Charles2017}. For lidar data, a frustum suffices to define an object in 3D space as 
lidar provides depth values. In our case, we only have 
data in the image space, without absolute depth value.
Instead of considering the whole frustum as in ~\cite{Charles2017}, we leverage the camera calibration and object dimensions in the training set to constrain the depth.
This idea is illustrated in Fig.~\ref{fig:depth_estimation}, 
where a person is located at different distances 
relative to the camera. Using the object height and 2D bounding box projection, we can estimate the distance to the camera through triangulation. Assuming a bounded height, we can accurately estimate the segment of the 
frustum where the object is located. In the example in Fig.~\ref{fig:depth_estimation}, 
we define the minimum and maximum height values to be 1.5m and 2m.

For each 2D bounding box $P=(c, u, v, w_{u}, h_{v})$ 
generated by the 2D detection network,
our 3D bounding box network is trained to estimate 
the parameters 
$Q'=(\delta u', \delta v', \delta z', \delta h', \delta w', \delta l', 
\theta')$, which encode the location $(x, y, z)$, dimensions $(h, w, l)$, and orientation $(\theta')$ of a 3D bounding box as follows


\vspace{-8pt}
\paragraph{3D Location.}
We estimate the objects location $(x, y, z)$ using 
its projection over the image space, 
as well as a frustum segment.
Specifically, we define the target $\delta u', \delta v'$ values 
as
\vspace{-6pt}
\begin{align} 
\delta u' &= (Proj2d_u (x,y,z) - u) / w_u \\ 
\delta v' &= (Proj2d_v (x,y,z) - v) / h_v, 
\vspace{-6pt}
\end{align}
where $Proj2d_u(x, y, z), Proj2d_v(x, y, z)$ represent the $u, v$ coordinates 
of the 2D projection of $(x, y, z)$ over the image space. 

To define the target $z$, we 
first define a frustum segment used as 
a reference for depth estimation.
Given an object with height $h$, 
we can estimate the object distance to the camera with focal length $f_v$ as
\vspace{-6pt}
\begin{equation}
\label{eq:depth}
f(h_{v}, h) = \frac{h}{h_{v}} f_v .
\vspace{-3pt}
\end{equation}
If we assume that $h$ 
follows a Gaussian Distribution with mean 
$\mu_h$ and standard deviation $\sigma_h$, 
given $P=(c, u, v, w_{u}, h_{v})$ and 
$f_v$, we can constrain 
the distance from the object to the camera to a range of
$ [f(h_{v}, \mu_h - \sigma_h), f(h_{v}, \mu_h + \sigma_h)]$, 
or, more general, we deduct that the frustum segment has a length $d$
\vspace{-4pt}
\begin{equation}
 d = f(h_{v}, \mu_h + k * \sigma_h) - f(h_{v}, \mu_h - k * \sigma_h),
\end{equation}
where $k$ is a scalar that adjusts the segment extent and
is inversely proportional to our prediction confidence. 

Following these observations, the z coordinate of the 3D bounding box, $\delta z'$, is given as
\vspace{-4pt}
\begin{equation} 
\delta z' = \frac{z - f(h_v, h)}{d}.
\vspace{-4pt}
\end{equation}
Note that learning $\delta z'$ instead of the absolute depth 
$z$ has the advantage that the target value includes a good depth estimation 
as prior and it is normalized by 
$d$, which varies according to the distance from the object to camera. 
We have found this normalization is key to estimate the absolute 
depth of the objects. Intuitively, for higher distances there is greater localization uncertainty in the labels and as such, the training loss needs to account for this proportionally. 
Analogous to 2D detectors, this frustum segment can also be considered as an anchor, except its position and dimensions are not fixed, instead using the camera model and object statistics to adjust accordingly.

During training, we use $h$ from ground-truth; during inference, 
we use the network prediction.


\vspace{-6pt}
\paragraph{3D Box Dimensions and Orientation.}
The target 3D box dimensions are estimated using  
$\delta h', \delta w', \delta l'$, which 
are defined as the offset between the mean of the objects dimensions, per class, and 
the \textit{true} dimensions.
\vspace{-2pt}
\begin{equation} 
\delta p' = \frac{p - \mu_p}{\mu_p}, \forall p \in \{h, w, l\}.
\vspace{-4pt}
\end{equation}
To learn the target orientation (observation angle) 
$\theta'$, the orientation is encoded as $(\sin_{\theta'}, \cos_{\theta'})$, 
and the network is trained to estimate each parameter separately.

\subsection{Loss Functions}
\label{sec:loss}
Given a 3D box parameters prediction 
$Q=(\delta u, \delta v, \delta z, \delta h, \delta w, \delta l, \sin_\theta, \cos_\theta)$, 
and its corresponding ground-truth box 
$Q'=(\delta u', \delta v', \delta z', \delta h', \delta w', \delta l', 
\theta')$, we define our overall loss 
$\mathcal{L}_\text{3D}(Q, Q')$ as
\vspace{-1pt}
\begin{equation}
\begin{aligned}
\mathcal{L}_\text{3D}(Q, Q') = \alpha * \sum_{l \in \{u, v, z\}} L_{loc}(\delta l - \delta l') \\
+ \sum_{d \in \{h, w, l\}} L_{dim} (\delta d - \delta d') +  \beta * L_{ori}(\sin_\theta, \cos_\theta, \theta'),
\end{aligned}
\vspace{-3pt}
\end{equation}
where  $L_{loc}$ is the location loss, $L_{dim}$ is the dimensions loss, 
and $L_{ori}(\theta, \theta')$ 
is the orientation loss. We use $\alpha$ and $\beta$ to weight the 
location and orientation loss, and define these values during training.
We define $L_{loc}$ and $L_{dim}$ as $SmoothL_1$, and $L_{ori}(\sin_\theta, \cos_\theta, \theta')$ as
\begin{equation}
 L_{ori}(\sin_\theta, \cos_\theta, \theta') = (\sin_\theta - \sin(\theta'))^2 + (\cos_\theta - \cos(\theta'))^2.
\end{equation}
The method runs at approximately 10~FPS on an Nvidia RTX 2080 GPU in TensorFlow without implementation optimization such as TensorRT. We refer to the Supplemental Material for additional method and implementation details. We also provide detailed ablation studies, validating the architecture components of the model, in the same document.

%
%

%
%

\section{Datasets}
\label{sec:dataset}
\begin{figure}[t!]
\centering
	\includegraphics[width=0.9\linewidth]{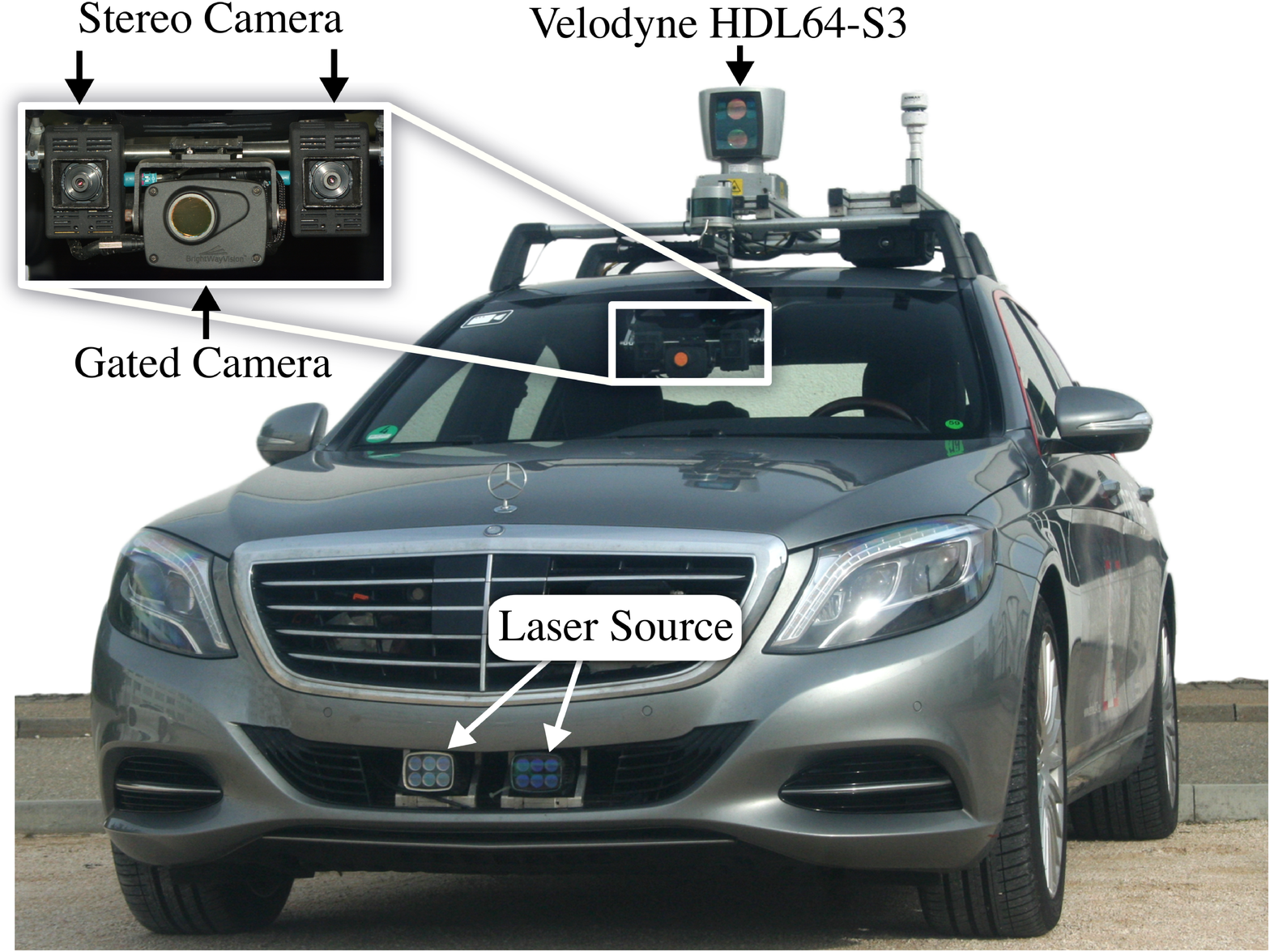}\\
\begin{minipage}[c]{0.9\linewidth}
	\resizebox{0.99\linewidth}{!}{
	\setlength{\tabcolsep}{0.5em}	
	{\renewcommand{\arraystretch}{1.5}
	\begin{tabular}{lccc}
			  & \textbf{Gated Camera} & \textbf{Stereo Camera} & \textbf{Lidar} \\[2pt]
			\toprule \addlinespace[5pt]
			\textbf{Sensor} & \makecell{BrightwayVision\\BrightEye} & \makecell{2x OnSemi\\AR0230} & \makecell{Velodyne\\HDL64-S3D} \\
			\textbf{Resolution} & 1280px$\times$720px & 1920px$\times$1080px & \arcsec{1440}$\times$\arcsec{612} \\
			\textbf{Wavelength} & \unit[808]{nm} & Color & \unit[905]{nm} \\
			\textbf{Frame Rate} & \unit[120]{Hz} & \unit[30]{Hz} & \unit[10]{Hz} \\
			\textbf{Bit Depth} & \unit[10]{bit} uint & \unit[12]{bit} uint & \unit[32]{bit} float \\ \addlinespace[2pt]
			\bottomrule
	\end{tabular}	
	}
	}
\end{minipage}
    \vspace*{-4pt}
	\caption{Sensor setup for recording the proposed Gated3D dataset. For comparisons we also capture corresponding lidar point clouds and stereo image pairs. Note that the stereo camera is located at approximately the same position of the gated camera in order to ensure a similar viewpoint.}
	\label{fig:sensor_setup}
\end{figure}

In this section, we describe \textit{Gated3D}, our new dataset for 
3D object detection with gated images.
\paragraph{Sensor Setup.}
Since existing automotive datasets \cite{waymo_open_dataset,cordts2016cityscapes,geiger2012we,yu2018bdd100k} do not include measurements from gated cameras, we collected gated image data during a large-scale data acquisition in Northern Europe. 
Following~\cite{gated2depth}, we used the gated system \emph{BrightEye} from BrightwayVision which consists of:
\begin{itemize}
	\item A gated CMOS pixel array of resolution $1280\text{ px} \times 720\text{ px}$ with a pixel pitch of \unit[10]{\textmu m}. Using a focal length of \unit[23]{mm} provides a horizontal and vertical field of view of $31.1^\circ \text{ H} \times 17.8^\circ \text{ V}$.
	\item Two repetitive pulsed vertical-cavity surface-emitting laser (VCSEL) which act as a pulsed illumination source. The VCSELs emit light at \unit[808]{nm} and \unit[500]{W} peak power to comply with eye-safety regulations. The pulsed illumination is diffused and results $24.0^\circ\text{ H} \times 8.0^\circ \text{ V}$ illumination cone. The source is mounted below the bumper of the vehicle, see Figure~\ref{fig:sensor_setup}.
\end{itemize}
The gated images consist of three exposure profiles as shown in Figure~\ref{fig:image_formation}. The corresponding gate settings (delay, laser duration, gate duration) can be found in the supplement. 
For each single capture, multiple laser flashes are integrated on the chip before read-out in order to increase the measurement signal-to-noise ratio.

For comparison with state-of-the-art 3D detection approaches, our test vehicle is equipped with a Velodyne HDL64 lidar scanner and a stereo camera. The stereo system consists of two cameras with OnSemi AR0230 sensors mounted at \unit[20.3]{cm} baseline. All sensor specifications are listed in Figure~\ref{fig:sensor_setup}.
The gated camera runs freely and cannot be triggered, so to obtain matching measurements we compensate the egomotion of the lidar point clouds. The corresponding gated images are found using an adapted ROS MessageFilter \cite{quigley2009ros}, see Supplemental Material.

\paragraph{Collection and Split}
We annotated 1.4 million frames collected at framerate of \unit[10]{Hz}, covering \unit[10,000]{km} of driving in Northern Europe during winter. The annotation and capture procedures for the dataset are detailed in the supplement. The gated images have been manually labeled with human annotators matching lidar, gated and RGB frames simultaneously. In total, more than 100,000 objects are labeled, which comprise 4 classes. The annotations were done over 12997 image examples. 
The dataset is randomly split into a training set of 10,046 frames, a validation set of 1,000 frames and a test set of 1,941 frames. 
In addition to the gated images, our proposed dataset contains corresponding RGB stereo images captured by the stereo camera system described in the previous paragraph.
%
%
In contrast to popular datasets, including as Waymo~\cite{waymo_open_dataset}, KITTI~\cite{geiger2012we} and Cityscapes~\cite{cordts2016cityscapes}, our dataset is significantly more challenging as it also includes many nighttime images and captures under adverse weather conditions such as snow and fog. 
%

\section{Assessment} 
\label{sec:experiments}

\begin{table*}[t!]
	\caption{Object detection performance over Gated3D dataset (test split). Our method outperforms monocular and stereo methods (bottom part of the table) over most of the short (0-30m), middle (30-50m) and long (50-80m) distance ranges, as well as 
	Pseudo-Lidar based methods trained over gated images. Interestingly, our model 
	even outperforms PointPillars lidar reference for Pedestrian detection at long distance ranges.}
	\centering
	\begin{subtable}{2.1\columnwidth}
	\setlength{\tabcolsep}{0.4em}
	\centering
	\vspace*{-4pt}
	\subcaption{Average Precision on \textit{Car} class.}
	\vspace*{-2pt}
	\resizebox{\columnwidth}{!}{
    \begin{tabular}{l|c|ccccccccc|ccccccccc}
		\multirow{3}{*}{\textbf{Method}} & \multirow{3}{*}{\textbf{Modality}} & \multicolumn{9}{c|}{\textbf{Daytime Images}} & \multicolumn{9}{c}{\textbf{Nighttime Images}} \\
		& & \multicolumn{3}{c}{\textbf{2D object detection}} & \multicolumn{3}{c}{\textbf{3D object detection}} & \multicolumn{3}{c|}{\textbf{BEV detection}} & \multicolumn{3}{c}{\textbf{2D object detection}} & \multicolumn{3}{c}{\textbf{3D object detection}} & \multicolumn{3}{c}{\textbf{BEV detection}} \\
		& & \unit[0-30]{m} & \unit[30-50]{m} & \unit[50-80]{m} & \unit[0-30]{m} & \unit[30-50]{m} & \unit[50-80]{m} & \unit[0-30]{m} & \unit[30-50]{m} & \unit[50-80]{m} & \unit[0-30]{m} & \unit[30-50]{m} & \unit[50-80]{m} & \unit[0-30]{m} & \unit[30-50]{m} & \unit[50-80]{m} & \unit[0-30]{m} & \unit[30-50]{m} & \unit[50-80]{m}\\
		\midrule
		\midrule



		\textsc{PointPillars} \cite{lang2019pointpillars} & Lidar & 90.12 & 82.83 & 56.63 & 91.51 & 84.63 & 54.28 & 91.59 & 86.54 & 54.71 & 90.73 & 84.88 & 54.22 & 90.29 & 87.40 & 52.32 & 90.29 & 87.51 & 52.60 \\
		\midrule
		\midrule

		\textsc{M3D-RPN} \cite{brazil2019m3d} & RGB & 90.44 & 89.29 & 62.76 & 53.21 & 13.26 & 10.52 & \textbf{60.80} & 16.16 & 10.52 & 90.85 & 80.64 & 59.76 & 51.18 & 20.76 & 2.73 & 52.53 & 21.39 & 2.74 \\

		\textsc{Stereo-RCNN} \cite{licvpr2019} & Stereo & 81.56 & 81.07 & 78.08 & \textbf{54.17} & 17.16 & 6.17 & 57.92 & 17.69 & 6.26 & 81.73 & 81.03 & 70.85 & 47.36 & 17.21 & 13.02 & \textbf{53.81} & 18.34 & 13.08 \\


		\textsc{Pseudo-Lidar} & Gated & 81.74 &  81.33 &  80.88 & 26.17 &  16.06 & 10.27 & 26.94  & 17.26 &  10.87 & 89.35 &  89.02 &  88.31 & 36.58  & 23.05 &  19.88 &
		39.50  & 28.68 &  22.82 \\

		\textsc{Pseudo-Lidar++} \cite{you2019pseudo} & Gated & 81.74  & 80.29 &  81.59 & 30.44 & 15.47 & 11.76 & 32.49 &  16.97 & 12.83 & 90.21 & 81.75 & 81.78 & 36.36 & 21.93   & \textbf{22.39} & 37.46 & 23.12 & \textbf{23.63} \\
		\textsc{PatchNet} \cite{ma2020rethinking} & Gated & 90.46 & 81.74 & 89.78 & 23.91 & 10.86 & 7.34 & 24.87 & 11.33 & 7.84 & \textbf{90.87} & \textbf{89.86} & 88.89 & 23.74 & 16.79 & 7.16 & 25.15 & 17.76 & 8.29 \\
		\textsc{Gated3D} & Gated & \textbf{90.78} & \textbf{90.55} & \textbf{90.91} & 52.15 & \textbf{28.31} & \textbf{14.85} & 52.31 & \textbf{29.26} & \textbf{15.02} & 90.84 & 81.82 & \textbf{90.33} & \textbf{51.42} & \textbf{25.73} & 12.97 & 53.37 & \textbf{29.13} & 13.12 \\
	\end{tabular}}
	\label{tab:3dgated_object_detection_results_car}
	\vspace*{4pt}
	\end{subtable}
\begin{subtable}{2.1\columnwidth}
	\setlength{\tabcolsep}{0.4em}
	\centering
	\vspace*{-4pt}
	\subcaption{Average Precision on \textit{Pedestrian} class.}
	\vspace*{-2pt}
	\resizebox{\columnwidth}{!}{
    \begin{tabular}{l|c|ccccccccc|ccccccccc}
		\multirow{3}{*}{\textbf{Method}} & \multirow{3}{*}{\textbf{Modality}} & \multicolumn{9}{c|}{\textbf{Daytime Images}} & \multicolumn{9}{c}{\textbf{Nighttime Images}} \\
		& & \multicolumn{3}{c}{\textbf{2D object detection}} & \multicolumn{3}{c}{\textbf{3D object detection}} & \multicolumn{3}{c|}{\textbf{BEV detection}} & \multicolumn{3}{c}{\textbf{2D object detection}} & \multicolumn{3}{c}{\textbf{3D object detection}} & \multicolumn{3}{c}{\textbf{BEV detection}} \\
		& & \unit[0-30]{m} & \unit[30-50]{m} & \unit[50-80]{m} & \unit[0-30]{m} & \unit[30-50]{m} & \unit[50-80]{m} & \unit[0-30]{m} & \unit[30-50]{m} & \unit[50-80]{m} & \unit[0-30]{m} & \unit[30-50]{m} & \unit[50-80]{m} & \unit[0-30]{m} & \unit[30-50]{m} & \unit[50-80]{m} & \unit[0-30]{m} & \unit[30-50]{m} & \unit[50-80]{m}\\
		\midrule
		\midrule


        \textsc{PointPillars} \cite{lang2019pointpillars} & Lidar & 70.08 & 49.03 & 0.00 & 69.71 & 45.24 & 0.00 & 70.53 & 48.07 & 0.00 & 69.97 & 43.32 & 0.00 & 71.25 & 41.21 & 0.00 & 70.99 & 43.61 & 0.00 \\
		\midrule
		\midrule

		\textsc{M3D-RPN} \cite{brazil2019m3d} & RGB & 79.08 & 66.41 & 36.98 & 26.20 & 14.50 & 9.84 & 30.68 & 17.47 & 10.07 & 78.36 & 62.99 & 36.76 & 25.09 & 6.43 & 2.07 & 26.42 & 7.69 & 2.74 \\

		\textsc{Stereo-RCNN} \cite{licvpr2019} & Stereo & 88.57 & 75.63 & 59.82 & 48.58 & \textbf{23.26} & 7.77 & 50.11 & \textbf{25.10} & 8.38 & 80.38 & 69.13 & 60.94 & 46.09 & 21.63 & 11.57 & 47.58 & 25.47 & 11.84 \\


		\textsc{Pseudo-Lidar} & Gated & 77.87 & 78.38 & 69.11 & 6.19 &  4.59 &  2.15 & 10.28 & 9.14  & 4.13 & 80.34 &  78.61 &  67.78 & 7.53  &  9.58  & 1.62 &
		14.27 &  15.72 &  5.55\\

		\textsc{Pseudo-Lidar++} \cite{you2019pseudo} & Gated & 77.89 &  77.95 & 60.88 & 9.19 & 2.36 & 3.30 & 14.32 &  5.66 &  4.10 & 79.84 & 79.57 & 54.42 & 7.37 & 7.21 & 2.06 & 12.92 & 11.99 & 5.64 \\
		\textsc{PatchNet} \cite{ma2020rethinking} & Gated & \textbf{90.48} & 80.75 & 69.56 & 32.88 & 18.05 & 5.62 & 39.45 & 20.27 & 9.77  & 81.50 & \textbf{88.62} & 65.43 & 15.37 & 13.37 & 6.75 & 21.60 & 18.15 & 8.46 \\
		\textsc{Gated3D} & Gated & 89.72 & \textbf{81.47} & \textbf{86.73} & \textbf{50.94} & 20.59 & \textbf{14.14} & \textbf{53.26} & 22.15 & \textbf{16.51} & \textbf{81.52} & 81.23 & \textbf{80.18} & \textbf{48.53} & \textbf{23.99} & \textbf{14.98} & \textbf{49.82} & \textbf{25.57} & \textbf{15.46} \\
	\end{tabular}}
	\label{tab:3dgated_object_detection_results_pedestrian}
	\vspace*{4pt}
	\end{subtable}

	\vspace*{-0pt}
	\label{tab:3dgated_object_detection_results}
\end{table*}

\paragraph{Evaluation Setting.}
The BEV and 2D/3D detection metrics as defined
in the KITTI evaluation framework are used for evaluation, as well as the ones
described by \cite{yang2018pixor}, which calculate the metrics with respect to distance ranges.
Following Simonelli et al.~\cite{simonelli2019disentangling},
average precision (AP) is based on 40 recall positions to provide a fair comparison.
We consider \emph{Pedestrian} and \emph{Car} as our target detection classes.

The 3D metrics are based on intersection
over union (IoU) between cuboids \cite{chen2017multi},
which has the disadvantage of equally penalizing completely
wrong detections and detections with IoU below the threshold. Due to the emphasis on challenging scenarios in the dataset, as well as imperfect sensor synchronization, the dataset has notably more label noise than typical public datasets for 3D object detection. This problem is mitigated
by using lower IoU thresholds than in KITTI: 0.2 for \emph{Car} and 0.1 for \emph{Pedestrian}. To focus on detection at different depth ranges, metrics based on difficulty, as defined in KITTI, are provided in the Supplemental Document.
\begin{figure*}[!t]
\includegraphics[trim={2.1cm 1.5cm 3.2cm 4.1cm},clip,width=\textwidth]{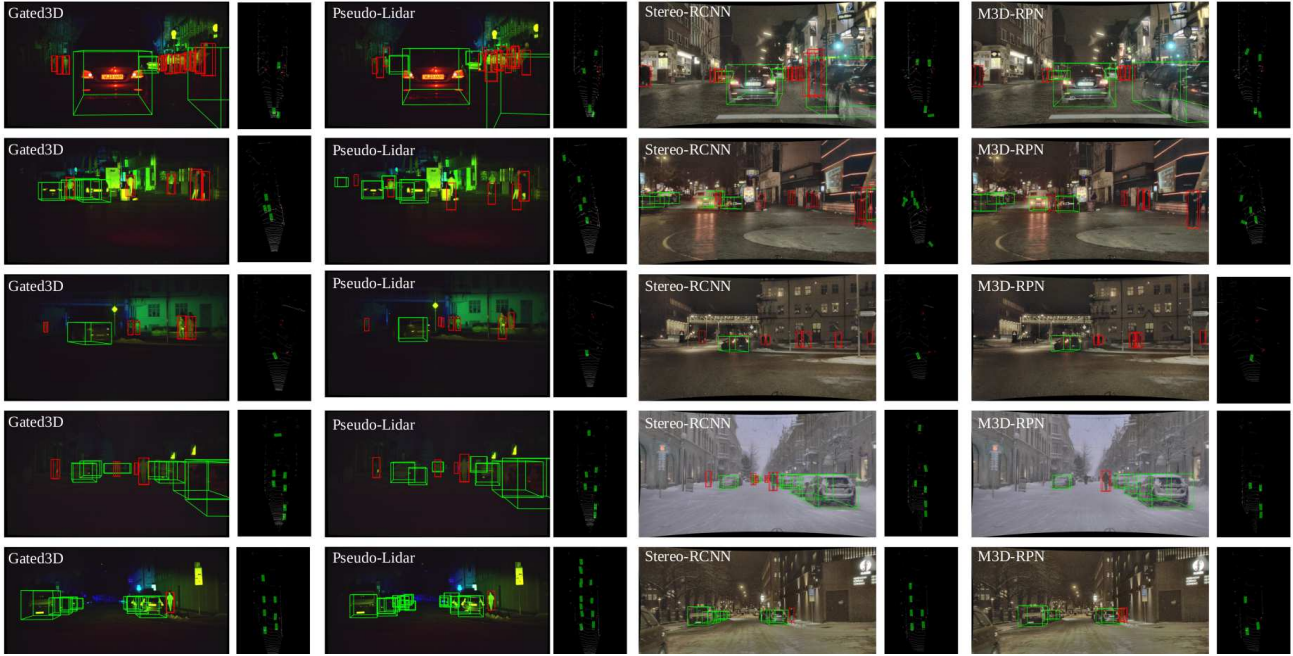}
\caption{Qualitative comparison against baseline methods on the captured dataset. Bounding boxes from the proposed method are tighter and more accurate than the state-of-the-art methods. This is seen in the second image with the other methods showing large errors in pedestrian bounding box heights. The BEV lidar overlays show our method offers more accurate depth and orientation than the baselines. For example, the car in the intersection of the fourth image has a 90 degree orientation error in the pseudo-lidar and stereo baselines, and is missed in the monocular baseline. The advantages of our method are most noticeable for pedestrians, as cars are easier for other methods due to being large and specular (please zoom in electronic version for details).}
\label{fig:real_comparison}
\end{figure*}

\paragraph{Baselines.}
We compare our approach to monocular,
stereo, lidar, and pseudo-lidar methods. As monocular 
baseline, we evaluate \textsc{M3D-RPN}~\cite{brazil2019m3d}, 
which performs 3D object detection from a single RGB image by 
``depth-aware'' convolution, where weights in one branch 
of the network are shared across rows only, assuming 
objects higher up in the image tend to be further away.
As stereo method, we evaluate \textsc{Stereo-RCNN}~\cite{licvpr2019}, 
which utilizes stereo image pairs to predict left-right 2D 
bounding boxes and keypoints that are then used to 
infer 3D bounding boxes using geometric constraints. 
Recent pseudo-lidar methods allow us to compare our method 
with recent state-of-the-art methods using the depth map as input, 
and therefore more directly asses the effectiveness 
of our model architecture in extracting information from 
gated images. To this end, we use the method from 
Gruber et. al.~\cite{gated2depth} 
to first generate dense depth maps from gated images, 
back-project all the pixels of the depth maps 
into 3D coordinates, and follow \cite{wang2019pseudo} 
to perform 3D object detection using Frustum PointNet~\cite{Charles2017}. 
We also evaluate Pseudo-Lidar ++ \cite{you2019pseudo} 
depth correction method from sparse lidar, downsampled 
from our 64 layered lidar to four lidar 
rays. Furthermore, we evaluate PatchNet \cite{ma2020rethinking}, 
which implements a pseudo-lidar approach based on 
2D image-based representation.
As a lidar reference method for reference with known (measured) depth, we evaluate 
\textsc{PointPillars} \cite{lang2019pointpillars}.

We use the corresponding open source repositories and tune the hyperparameters
of each baseline model during training over our dataset.

\paragraph{Experimental Validation.}
%
Tables~\ref{tab:3dgated_object_detection_results_car} and ~\ref{tab:3dgated_object_detection_results_pedestrian}, respectively, show \emph{Car} and \emph{Pedestrian} AP for 2D, 3D and BEV detection on the test set. These results demonstrate the utility of gated imaging for 3D object detection. Consistent with prior work ~\cite{licvpr2019} both the monocular and stereo baselines show a drop in performance with increasing distance. Monocular and stereo depth cues for a small automotive baseline of 10 - 30cm are challenging to find with increasing range.

The proposed \textsc{Gated3D} method offers a new image modality between monocular, stereo and lidar measurements. The results demonstrate improvement over intensity-only methods, especially for pedestrians and at night. \textsc{Gated3D} excels at detecting objects at long distances or in low-visibility situations. Note that pseudo-lidar and stereo methods can be readily combined with the proposed method --- a gated stereo pair may capture stereo cues orthogonal to the gated cues exploited by the proposed method. 
For additional ablation studies on the components of the proposed method, please refer to the Supplemental Document.

%


Figure~\ref{fig:real_comparison}
shows qualitative examples of our proposed method
and state-of-the-art methods. 
The color-coded gated images
illustrate the semantic and space information
of the gated data (red tones for closer
objects and blue for farther away ones).
Our method accurately detects objects
at both close and large distances,
whereas other methods struggle, particularly in the safety-critical application of detecting pedestrians at night or in adverse weather conditions.




\section{Conclusions and Future Work}
This work presented the first 3D object detection method for gated images. As a low-cost alternative to lidar, \emph{Gated3D} outperforms recent stereo and monocular detection methods, including state-of-the-art pseudo-lidar approaches. We expand on CMOS sensor arrays used in passive imaging approaches by flood-illuminating the scene and capture the temporal intensity variation in coarse temporal gates. Gated images allow us to leverage existing 2D feature-extraction architectures. We distribute the resulting features in the camera frustum along the corresponding gate -- a representation that naturally encodes geometric constraints between the gates, without the need to first recover intermediate proxy depth maps. The proposed method runs at real-time rates and we validate the method experimentally on 10,000~km of driving data, demonstrating \emph{higher 3D object detection accuracy than existing monocular or stereo detection methods}, including recent stereo and monocular pseudo-lidar methods with similar cost to the proposed system. The proposed method allows for accurate object detection in low-illumination scenarios, where passive methods fail, while being a low-cost camera with an additional flash source.

In the future, gated imaging systems could benefit from stereo cues (in a stereo system). We envision our work as a first step towards gated imaging as a new sensor modality, beyond lidar, radar and camera, useful for a broad range of tasks in robotics and autonomous driving, including tracking, motion planning, SLAM, visual odometry, and large-scale scene understanding.

\clearpage
{\small
\bibliographystyle{ieee}
\bibliography{bib}
}

\end{document}